\documentclass{article} % For LaTeX2e
\usepackage[T1]{fontenc}
\usepackage{iclr2027_conference,times}

\usepackage{hyperref}
\hypersetup{hidelinks}
\usepackage{url}
\usepackage{wrapfig}
\usepackage{microtype}

\usepackage{graphicx}
\usepackage{array}
\usepackage{amsmath}
\usepackage{bm}
\usepackage{amsmath}
\usepackage{multirow}
\usepackage{indentfirst}
\usepackage{graphicx}
\usepackage{url}
\usepackage{xspace}
\usepackage{booktabs}
\usepackage{color}
\usepackage{amssymb}

\newcommand{\paratitle}[1]{\vspace{1.5ex}\noindent\textbf{#1}}

\newcommand{\ignore}[1]{}

\title{From Dissonance to Orchestration:
Teacher\\Intervention in On-Policy Distillation}

\author{
Yuhao Wang$^{1}$,
Ruiyang Ren$^{1}$\thanks{Corresponding authors.},
Yinan Zhang$^{1}$\footnotemark[1],
Ruiqing Zhang$^{2}$,
Jing Liu$^{2}$,
Chunyan Miao$^{1}$ \\
$^{1}$Nanyang Technological University, Singapore \\
$^{2}$Baidu Inc. \\
\texttt{yh.wang500@outlook.com}
}

\iclrfinalcopy
\begin{document}

\maketitle

\begin{abstract}
On-policy distillation (OPD) trains a student on its own reasoning trajectories using feedback from a stronger teacher. Teacher interventions can improve these trajectories, but also change the distribution on which the student learns. Our controlled studies show that rollout quality alone is an incomplete criterion for allocating teacher guidance. Deeper intervention yields diminishing gains in rollout accuracy while increasing off-policy load. In a training probe with a restricted rollout horizon, peak student accuracy and performance retention favor different intervention strengths. The preferred intervention depth and placement also vary across benchmarks. These findings motivate \textbf{MAESTRO}, which uses local policy disagreement to jointly adapt when the teacher takes over and how long it generates. Its {policy disagreement score} combines teacher-weighted candidate coverage with local distribution similarity and is aggregated within reasoning paragraphs. Across eight mathematical reasoning benchmarks, MAESTRO achieves the highest macro-average accuracy among the compared methods for both 0.6B and 1.7B Qwen3 students, with the 1.7B student leading on every benchmark. MAESTRO also reduces average training response length by 67.3\% relative to standard OPD. The code is available at \url{https://github.com/yhao-wang/MAESTRO}.

\end{abstract}

\section{Introduction}
\label{sec:introduction}

On-policy distillation (OPD)~\citep{DBLP:conf/iclr/AgarwalVZSGGB24} trains a language
model on responses sampled from the student's current policy. The teacher
provides token-level supervision on student-generated prefixes, reducing the
distribution mismatch caused by teacher-generated training sequences. For reasoning, this alignment is particularly important because each
intermediate step becomes part of the prefix for subsequent generation.
At the same time, it ties the training distribution to the student's current capabilities: errors in student-generated prefixes directly shape the states on which later supervision is provided~\citep{DBLP:journals/jmlr/RossGB11,DBLP:conf/nips/BengioVJS15}.

An early reasoning error can lead to a long, unproductive continuation along which teacher feedback becomes less useful~\citep{DBLP:conf/iclr/0009CMZYSZ24,zhou2026less}. Temporarily handing generation to the teacher can redirect such trajectories~\citep{DBLP:conf/iclr/0006YZXG0Y25}. However, its influence persists after control returns: the student resumes from a prefix partly generated by the teacher, which shapes its subsequent reasoning and supervision. Intervention thus changes both the quality of the generated response and the distribution of prefixes on which the student learns. 
We view this tension between corrective teacher guidance and departure from the student's own policy as a form of policy dissonance.

\begin{figure*}[t]
    \centering
    \includegraphics[width=\textwidth]{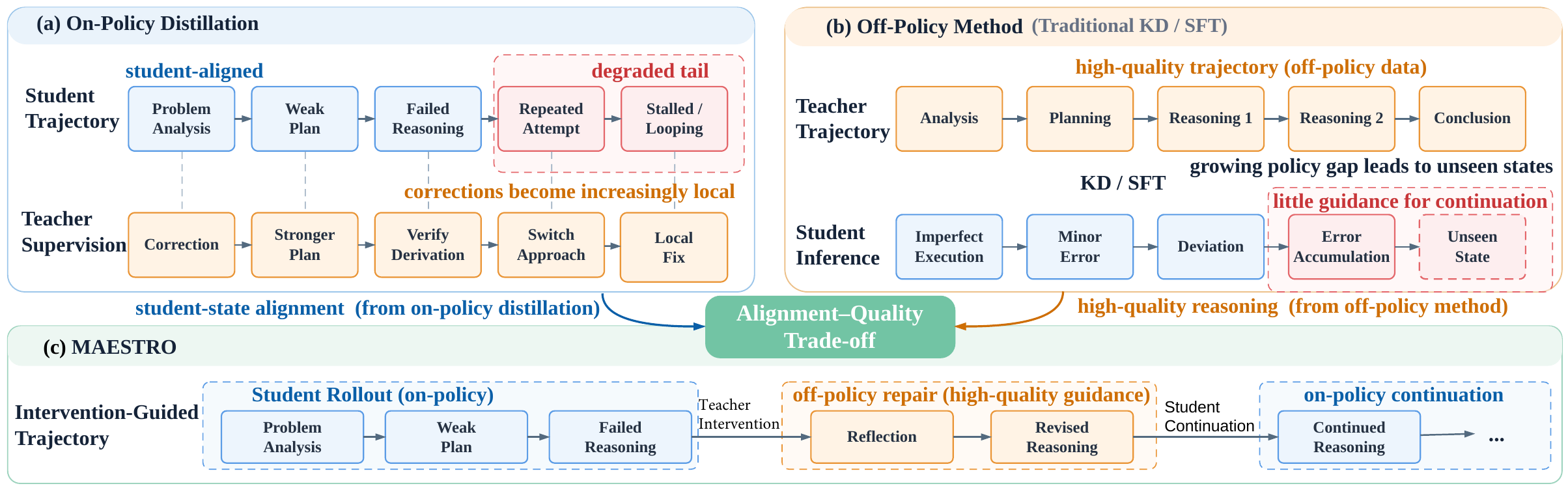}
    \caption{
    The core motivation of MAESTRO is to balance the benefit of teacher correction
against the off-policy shift introduced by teacher generation.
% MAESTRO uses
% local policy disagreement to adapt when the teacher intervenes and how deeply
% it takes over.
    }
    \label{fig:motivation}
    \vspace{-10pt}
\end{figure*}

Speculative knowledge distillation replaces selected student-proposed tokens according to the teacher's distribution~\citep{DBLP:conf/iclr/XuH0LM0WALP25,DBLP:conf/acl/ZhangLSYG025}, while Relay-OPD uses a reflection-based trigger to initiate teacher spans with a prescribed number of paragraphs~\citep{DBLP:journals/corr/abs-2607-26057}. These methods establish the value of teacher participation during rollout. We use controlled interventions to examine how the amount and placement of teacher generation affect both rollout quality and the accuracy attained and retained by the student during training.

Our analysis distinguishes the effects of teacher intervention on trajectory quality from its effects on student learning. Deeper intervention improves rollout accuracy with diminishing returns, while increasing off-policy load. We quantify this load using both the fraction of teacher-generated tokens and the log-probability gap between the two policies on those tokens; intervention rate alone does not describe their deviation from the student policy. In a training probe with a restricted rollout horizon, moderate intervention yields the highest peak student accuracy, whereas stronger intervention produces a lower peak but preserves a larger fraction of it by the end of training. Reaching a high peak and retaining performance therefore favor different intervention strengths. Rollout quality alone is thus an incomplete criterion for deciding how much teacher guidance to introduce.

The preferred intervention depth and relative position also vary across benchmarks.
More challenging benchmarks favor deeper teacher intervention, while the earliest
tested position is not always the most effective. These findings motivate a more
adaptive orchestration of teacher intervention, in which both timing and depth
respond to the current reasoning state. Local policy disagreement provides an
observable signal for this adaptation, reflecting how the student's continuation
diverges from the teacher's as reasoning unfolds and the student policy evolves.

We introduce \textbf{MAESTRO}, which uses paragraph-level policy feedback to
adapt both intervention timing and depth. Its \emph{Policy Disagreement Score}
(PDS) combines two signals: the student's coverage of teacher-preferred
candidates and the similarity of their relative probability assignments,
with candidate coverage weighted by teacher probability.
MAESTRO aggregates token-level PDS within each reasoning paragraph and
evaluates the resulting signal at paragraph boundaries. A score above the
threshold triggers teacher intervention, and larger disagreement leads to
deeper intervention. Each rollout allows a limited number of
interventions, with student
generation resuming between them.

We evaluate MAESTRO on eight mathematical reasoning benchmarks with 0.6B and 1.7B Qwen3 students. Relative to Relay-OPD, MAESTRO improves macro-average accuracy from 29.71\% to 31.36\% for the 0.6B student and from 45.18\% to 47.12\% for the 1.7B student. With the 1.7B student, it achieves the highest accuracy among the compared methods on all eight benchmarks. MAESTRO also reduces average training response length by 67.3\% relative to OPD. Ablations support the contributions of the two score components, weighted aggregation, and adaptive intervention duration. Teacher usage also varies across benchmarks and over training: MAESTRO uses more teacher tokens than Relay-OPD on the more challenging benchmarks, while the fraction of teacher-generated tokens in its rollouts decreases over training.

Our contributions are:
\begin{itemize}
    \item We characterize teacher intervention in OPD across trajectory
behavior, training dynamics, and adaptive allocation, revealing near-logarithmic
scaling, improved performance and stability, and difficulty- and progress-aware
adaptation of intervention depth and position.
    \item We introduce MAESTRO, which combines complementary measures of policy disagreement into paragraph-level feedback for jointly adapting intervention timing and depth.
    \item We demonstrate higher macro-average accuracy on eight mathematical reasoning benchmarks at two student scales, together with shorter training responses relative to OPD and ablations supporting the intervention design.
\end{itemize}

\section{Characterizing Teacher Intervention in OPD}
\label{sec:analysis}

In this section, we characterize how teacher intervention affects OPD.
Specifically, we study the trade-off between preserving on-policy student
rollouts and introducing off-policy teacher guidance, and examine how this
trade-off manifests in trajectory behavior, training dynamics, and
intervention allocation. We organize our analysis around three questions:
\textbf{RQ1:} What trade-off emerges as teacher intervention increases?
\textbf{RQ2:} How does teacher intervention reshape training effectiveness and efficiency?
\textbf{RQ3:} How should teacher intervention be allocated across reasoning trajectories?
We address these questions, respectively, from the trajectory, training, and adaptive perspectives.

\begin{figure*}[t]
    \centering
    \includegraphics[width=\textwidth]{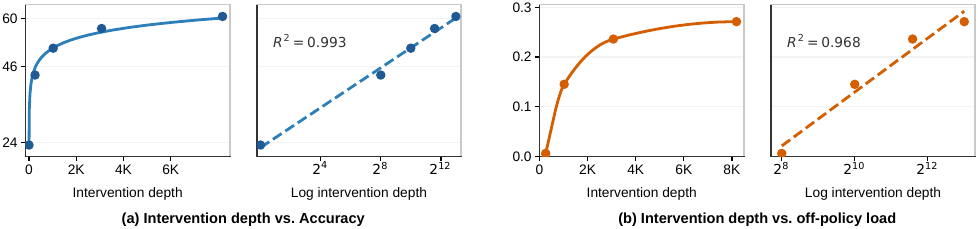}
    \caption{
{Scaling trade-off of teacher intervention.}
We vary teacher intervention depth and measure its effect on
{(a)} trajectory accuracy and {(b)} trajectory-normalized
off-policy load. 
}
    \label{fig:intervention_scaling}
\end{figure*}

\subsection{Scaling Trade-off of Teacher Intervention}

To understand what trade-off emerges as teacher intervention increases, we examine intervention from the trajectory perspective.
Specifically, we vary the intervention depth and examine how increasing teacher guidance affects trajectory quality and off-policy deviation.
The results are shown in Figure~\ref{fig:intervention_scaling}.

\paratitle{Intervention and Trajectory Quality.}
We first study how trajectory quality changes with intervention depth.
As shown in the left panel of Figure~\ref{fig:intervention_scaling}(a),
accuracy increases continuously as the teacher intervenes more deeply, with
larger gains at shallow depths and progressively smaller gains thereafter.
This pattern suggests a logarithmic scaling relationship. When intervention
depth is shown on a log scale in the right panel of
Figure~\ref{fig:intervention_scaling}(a), the points become nearly linear,
with $R^2=0.993$, confirming the approximately logarithmic trend.

\paratitle{Intervention and Off-Policy Deviation.}
We next study how off-policy deviation changes with intervention depth.
Teacher-token ratio alone cannot fully characterize the resulting off-policy
deviation, since the likelihood of a teacher-generated token depends on both
the current prefix and the student policy.
Building on prior work that uses the log-probability difference between the
teacher and student to characterize token-level policy discrepancy~\citep{DBLP:journals/corr/abs-2603-22117}, we aggregate this quantity over teacher-generated
positions and normalize it by the trajectory length:
\begin{equation}
    L_{\mathrm{off}}
    =
    \frac{1}{N}
    \sum_{t\in T}
    \left[
        \log \pi_T(a_t^T \mid s_t)
        -
        \log \pi_S(a_t^T \mid s_t)
    \right],
\end{equation}
where $T$ denotes the set of teacher-generated positions and $N$ is the
response length. This definition can be decomposed into the amount of teacher intervention and
the policy discrepancy on teacher-generated tokens:
\begin{equation}
    L_{\mathrm{off}}
    =
    \rho_T
    \log
    \frac{\mathrm{PPL}_S^T}
         {\mathrm{PPL}_T^T},
    \qquad
    \rho_T=\frac{|T|}{N}.
\end{equation}
Here, $\rho_T$ captures how much of the trajectory is generated by the
teacher, while the log-perplexity ratio measures how unlikely these tokens are
under the student relative to the teacher given the corresponding prefixes.
As shown in the left panel of Figure~\ref{fig:intervention_scaling}(b),
$L_{\mathrm{off}}$ increases continuously with intervention depth. The
increase is steepest at shallow depths and becomes progressively slower as
intervention deepens, indicating that even limited teacher intervention can
introduce substantial off-policy deviation. When intervention depth is shown on a log scale in the right panel, the
relationship is well fitted by a logarithmic function, with $R^2=0.968$.

\paratitle{Finding 1.
A small amount of teacher intervention is already sufficient to recover much
of the trajectory quality, but it also incurs a substantial off-policy cost.}
The two approximately logarithmic trends show that both the benefit and the
cost are concentrated at shallow intervention depths, making selective
intervention crucial rather than simply increasing teacher involvement.

\begin{figure*}[t]
    \centering
    \includegraphics[width=\textwidth]{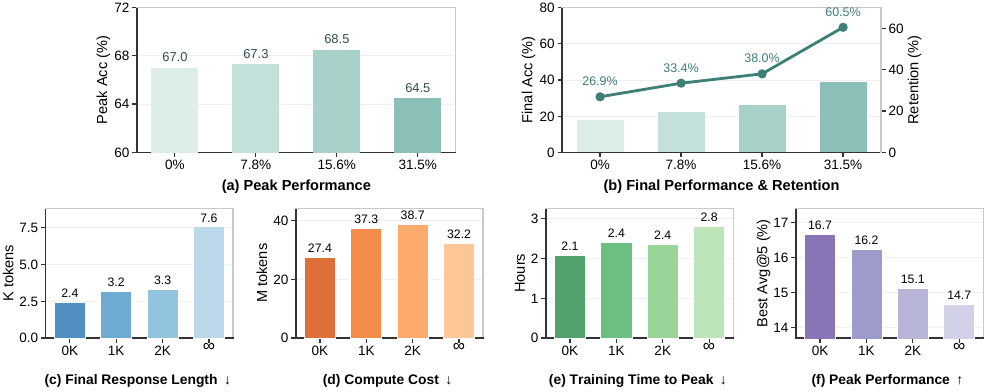}
\caption{
{Teacher intervention under a limited rollout length.}
{(a--b)} We vary intervention strength to examine peak performance and
training stability.
{(c--f)} We fix the intervention schedule and vary post-intervention
student continuation to examine training efficiency.
}
    \label{fig:training_tradeoff}
    \vspace{-3pt}
\end{figure*}

\subsection{Training Trade-off of Teacher Intervention}

To understand how teacher intervention reshapes training effectiveness and
efficiency, we turn to the training perspective. Specifically, we conduct two
complementary analyses. First, we fix the maximum rollout length and vary the
amount of teacher intervention to examine its effects on attainable performance
and training stability. Second, we fix the intervention schedule and vary the
post-intervention student continuation to examine whether teacher guidance can
reduce the amount of rollout required for effective training.

\subsubsection{Intervention under Limited Rollout Length}

To isolate the effect of teacher intervention, we keep the maximum rollout
length fixed and vary only the intervention amount. The results are shown in
Figure~\ref{fig:training_tradeoff}(a--b).

\paratitle{Intervention Raises the Performance Peak.}
As shown in Figure~\ref{fig:training_tradeoff}(a), teacher intervention leads
to two observations. First, it can raise the attainable performance: the
15.6\% setting achieves the highest peak accuracy of 68.5\%, while both 7.8\%
and 15.6\% outperform pure OPD at 67.0\%. Second, the intervention level must
be carefully controlled. With 31.5\% intervention, the peak accuracy is only
64.5\%, lower than that of pure OPD.

\paratitle{Intervention Extends Training Stability.}
Figure~\ref{fig:training_tradeoff}(b) shows that teacher intervention
consistently improves training stability. All intervention settings retain a
larger fraction of their peak performance than pure OPD. This effect becomes
stronger with more intervention, with the 31.5\% setting retaining about
2.25$\times$ as much peak performance as pure OPD. Thus, although stronger
intervention does not further improve the performance peak, it can
substantially mitigate the degradation that follows.

\subsubsection{Reducing Rollout Cost with Teacher Intervention}
\label{sec:analysis-tail}

We next examine whether teacher intervention can reduce the amount of student
rollout required during training. We fix the intervention schedule and vary only
the post-intervention student continuation budget among 0, 1K, 2K, and 4K
tokens. The results are shown in
Figure~\ref{fig:training_tradeoff}(c--f).

\paratitle{Shorter Rollouts Improve Training Stability.}
As shown in Figure~\ref{fig:training_tradeoff}(c,f), even with only about
2.4K rollout tokens, the shortest continuation setting still reaches the
highest performance of 16.7\%. This suggests that, once teacher intervention
has corrected the trajectory, long student continuation is not necessary for
maintaining strong training performance. Shorter continuation also increases
the relative share of teacher-generated tokens in the trajectory while
reducing unnecessary student-side rollout. This observation is consistent with
Figure~\ref{fig:training_tradeoff}(b), where a higher proportion of teacher
guidance leads to stronger performance retention and more stable training
after the peak.

\paratitle{Shorter Rollouts Improve Training Efficiency.}
Figure~\ref{fig:training_tradeoff}(d--e) shows that shorter continuation
reduces both rollout cost and wall-clock training time. The shortest
continuation setting reaches its peak in only 2.1 hours, compared with
2.8 hours under unrestricted continuation. Notably, the reduction in training
time is more pronounced than that in rollout tokens, since generation is often
bottlenecked by the longest trajectories in a batch. Limiting
post-intervention continuation therefore reduces both generation cost and the
straggler effect from long rollouts. This highlights another benefit of
teacher intervention: once the trajectory is corrected, less student rollout
is required to achieve strong performance.

\paratitle{Finding 2.
Teacher intervention improves the training trade-off from both performance and
efficiency perspectives.} Moderate intervention raises the attainable
performance, while stronger intervention improves training stability. This
stabilizing effect further reduces the need for long student rollouts,
allowing strong performance to be reached with lower rollout cost.

\begin{wrapfigure}[14]{R}{0.40\textwidth}
    \vspace{-0.8\baselineskip}
    \centering
    \includegraphics[width=\linewidth]{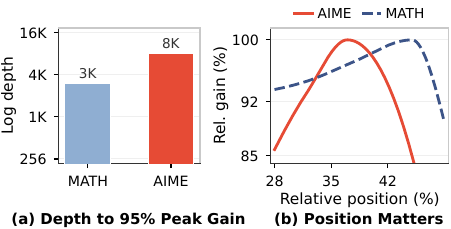}
    \caption{
Adaptive allocation of teacher intervention.
{(a)} Harder problems require greater intervention depth.
{(b)} The preferred intervention position also varies with reasoning
progress.
}
    \label{fig:adaptive_allocation}
    % \vspace{-0.6\baselineskip}
\end{wrapfigure}

\subsection{Adaptive Allocation of Teacher Intervention}

To understand how teacher intervention should be allocated across reasoning
trajectories, we turn to the adaptive perspective. Specifically, we vary intervention depth and relative position to examine how
teacher guidance should adapt to problem difficulty and reasoning progress. The
results are shown in Figure~\ref{fig:adaptive_allocation}.

\paratitle{Harder Problems Require Deeper Intervention.}
We first examine how much teacher intervention is needed across problems of
different difficulty. As shown in Figure~\ref{fig:adaptive_allocation}(a),
MATH reaches 95\% of its peak gain with about 3K intervention tokens,
whereas AIME requires about 8K. This indicates that harder problems require
substantially deeper teacher intervention to obtain a comparable fraction of
their attainable improvement.

\paratitle{Effective Intervention Depends on Relative Position.}
We next study where teacher intervention should occur along the reasoning
trajectory. As shown in Figure~\ref{fig:adaptive_allocation}(b), performance
is highest when intervention is introduced at an intermediate relative
position rather than too early or too late. The preferred position also differs
across datasets, indicating that a fixed intervention point cannot generalize
across trajectories with different reasoning progress.

\paratitle{Finding 3.
Teacher intervention should adapt to problem difficulty in both depth and
position.} Harder problems require deeper intervention, while the preferred
intervention point changes with reasoning progress. This makes a fixed
intervention strategy suboptimal across problems of different difficulty.

\section{MAESTRO}
\label{sec:maestro}

The preceding analysis shows that teacher intervention affects both rollout quality and the distribution on which the student learns, while its benefits depend on intervention depth and placement. We introduce \textbf{MAESTRO}, which uses feedback from the student and teacher policies to allocate teacher generation as a rollout unfolds. Its \textit{Policy Disagreement Score} (PDS) measures local differences in their continuation preferences and guides two coupled decisions: when the teacher takes over and how much of the trajectory it generates. We first define PDS, then describe the intervention rule.

\subsection{Policy Disagreement Score}
\label{sec:pds}

Candidate overlap alone cannot reveal how differently the student and teacher weight shared tokens. Conversely, similarity within the teacher's top-$K$ region can remain high after normalization even when those tokens fall outside the student's leading choices. These two limitations motivate PDS as a local measure of policy disagreement.

At each position $t$, PDS combines two complementary signals: teacher-mass
coverage captures whether the student retains the teacher's preferred
candidates, while local distributional similarity measures whether the two
policies assign similar preferences within this region. We denote the teacher
and student top-$K$ token sets as $T_K(t)$ and $S_K(t)$, respectively. The
teacher-mass coverage is defined as
\begin{equation}
C_t =
\sum_{v\in T_K(t)}
\tilde{p}_T(v)\,
\mathbf{1}[v\in S_K(t)],
\end{equation}
where $\tilde{p}_T$ denotes the teacher probability normalized over its top-$K$
tokens. Unlike binary top-$K$ overlap, $C_t$ weights each matched candidate by
its teacher probability, assigning greater importance to tokens preferred more
strongly by the teacher.

We further measure the local distributional similarity using the Bhattacharyya
coefficient~\citep{bhattacharyya1943measure}:
\begin{equation}
B_t =
\sum_{v\in T_K(t)}
\sqrt{\tilde{p}_T(v)\tilde{p}_S(v)},
\end{equation}
where $\tilde{p}_S$ denotes the student probability normalized over the
teacher's top-$K$ region. $B_t$ therefore captures the similarity of relative
probability assignments within the teacher's high-probability region.

We combine the two complementary signals to obtain the token-level PDS:
\begin{equation}
\mathrm{PDS}_t = 1 - C_t B_t.
\end{equation}
By taking their product, PDS requires agreement in both candidate coverage and
local probability assignment. A mismatch in either aspect therefore increases
PDS and cannot be compensated for by strong agreement in the other. As a
result, a larger PDS indicates stronger local disagreement between the student
and teacher policies.

\paratitle{Paragraph-level Aggregation.}
Token-level PDS captures policy disagreement only at individual positions,
which may not reliably reflect a persistent policy deviation. Directly relying
on such point-wise signals can therefore lead to spurious intervention
decisions. We aggregate token-level PDS within each reasoning paragraph to
obtain a more stable estimate:
\begin{equation}
\overline{\mathrm{PDS}}_p =
\frac{
\sum_{i=0}^{L_p-1} w_i\,\mathrm{PDS}_{p,i}
}{
\sum_{i=0}^{L_p-1} w_i
},
\qquad
w_i = 1+\alpha e^{-i/\tau},
\end{equation}
where $L_p$ denotes the number of tokens in paragraph $p$.
We assign larger weights to earlier positions in the paragraph, as the early
part of a reasoning segment often determines its direction, while later tokens
may simply continue along an established path. Disagreement at earlier
positions can therefore be more informative for identifying potential
deviations. This weighted aggregation also reduces the effect of isolated
token-level variations, providing a more stable signal for teacher
intervention.

\subsection{Adaptive Teacher Intervention}
\label{sec:adaptive-intervention}

\paratitle{Dynamic intervention.}
A fixed intervention strategy applies the same teacher guidance regardless of
the current policy disagreement. This can introduce unnecessary off-policy
guidance when the student remains close to the teacher, while providing
insufficient correction when the two policies differ substantially. We
therefore use PDS to adapt teacher intervention to the current policy gap.
Specifically, MAESTRO controls two decisions: \emph{when} the teacher should
take over and \emph{how deep} the intervention should go.

\paratitle{Intervention Timing.}
MAESTRO uses the aggregated PDS to determine when teacher intervention is
needed. At each paragraph boundary, we evaluate
$\overline{\mathrm{PDS}}_p$. When the score exceeds a threshold $\delta$,
teacher intervention is triggered. This allows the student to continue its
own rollout when the two policies remain close, while introducing teacher
guidance when their disagreement becomes large.

\paratitle{Intervention Depth.}
Beyond deciding when to intervene, the aggregated PDS also controls
intervention depth. Once the intervention threshold is crossed, MAESTRO
increases the number of teacher-generated paragraphs with
$\overline{\mathrm{PDS}}_p$. This provides stronger correction for severely
deviated trajectories, while limiting unnecessary teacher generation when the
policy gap is small.

\subsection{Discussion}

MAESTRO is designed to translate the findings in Section~\ref{sec:analysis}
into concrete intervention decisions. Its PDS-based trigger makes teacher
guidance selective rather than uniform, reflecting the RQ1 trade-off between
trajectory improvement and off-policy deviation. Its rollout pipeline terminates
after the final intervention, following the RQ2 observation that effective
correction can reduce unnecessary student continuation and rollout cost.
Finally, paragraph-level triggering together with disagreement-dependent
intervention depth realizes the RQ3 finding that useful teacher guidance should
adapt to both reasoning progress and problem difficulty. Together, these choices
shift teacher intervention from a fixed amount of assistance to a
state-dependent allocation mechanism.
\section{Experiment}

\subsection{Experimental Setup}

\paratitle{Training Data and Evaluation.}
We use the English subset of DAPO-Math-17K as the training data~\citep{DBLP:conf/nips/YuZZYZYDFLLLLLL25}.
We evaluate all methods on eight mathematical reasoning benchmarks:
AIME 2024, AIME 2025, AIME 2026, MATH500~\citep{DBLP:conf/iclr/LightmanKBEBLLS24},
AMC 2023, OlympiadBench~\citep{DBLP:conf/acl/HeLBHTSHHHZLQL024},
HMMT February 2026, and HMMT November 2025~\citep{DBLP:journals/corr/abs-2605-00674}.
We report avg@32 accuracy for all benchmarks.

\paratitle{Models.}
We use Qwen3-4B-Instruct-2507~\citep{DBLP:journals/corr/abs-2505-09388} as the teacher model and conduct experiments
with two student models, Qwen3-0.6B-Non-Thinking and
Qwen3-1.7B-Non-Thinking~\citep{DBLP:journals/corr/abs-2505-09388}.
This setup allows us to evaluate MAESTRO across students with different
model capacities.

\paratitle{Baselines.}
We compare MAESTRO with the initial student, SFT, KD
~\citep{DBLP:journals/corr/HintonVD15},
GRPO~\citep{DBLP:journals/corr/abs-2402-03300},
OPD~\citep{DBLP:conf/iclr/AgarwalVZSGGB24},
TRD~\citep{DBLP:journals/corr/abs-2606-08432},
FastOPD~\citep{DBLP:conf/acl/ZhangYJHRQLBT26},
SKD~\citep{DBLP:conf/iclr/XuH0LM0WALP25},
and Relay-OPD~\citep{DBLP:journals/corr/abs-2607-26057}.
These baselines cover supervised distillation, reinforcement learning,
standard OPD, and different trajectory-level intervention
strategies.

\paratitle{Implementation Details.}
All online distillation methods are trained for one epoch with a maximum
response length of 8192 tokens. Trajectory sampling uses a temperature of
1.0 and top-$p$ of 1.0. For MAESTRO, we allow at most two teacher
interventions per trajectory. The rollout is terminated after the final
intervention. Unless otherwise specified, we use the same MAESTRO
configuration for both student models and across all evaluation benchmarks.

\begin{table*}[t]
\centering
\caption{
Main results on eight mathematical reasoning benchmarks.
We report mean accuracy (\%) on each benchmark and the macro-average across all benchmarks.
}
\label{tab:main_results}

{
\small
\setlength{\tabcolsep}{3.5pt}
\renewcommand{\arraystretch}{1.05}

\begin{tabular}{lccccccccc}
\toprule
Method
& AIME24 & AIME25 & AIME26
& AMC23 & HMMT26 & HMMT25
& MATH500 & Olymp. & Avg. \\
\midrule

\multicolumn{10}{c}{\textit{Qwen3-0.6B-Non-Thinking}} \\
\midrule

Student
& 1.77 & 2.40 & 0.73
& 24.45 & 0.76 & 3.85
& 44.10 & 16.36 & 11.80 \\

SFT
& 4.90 & 7.60 & 4.06
& 34.92 & 1.89 & 3.02
& 59.45 & 26.74 & 17.82 \\

KD
& 4.17 & 7.19 & 4.79
& 35.23 & 2.94 & 2.71
& 57.75 & 27.15 & 17.74 \\

GRPO
& 8.23 & 15.21 & 10.00
& 46.56 & 7.86 & 6.04
& 68.60 & 35.42 & 24.74 \\

TRD
& 4.58 & 8.54 & 3.96
& 36.33 & 3.79 & 2.81
& 57.60 & 26.93 & 18.07 \\

SKD
& 8.44 & 16.98 & 9.79
& 43.67 & 8.14 & 5.62
& 66.65 & 35.76 & 24.38 \\

OPD
& 15.31 & 16.67 & 10.00
& 42.58 & 5.49 & 6.04
& 74.30 & 46.89 & 27.16 \\

FastOPD
& \textbf{18.65} & \underline{17.29} & 11.98
& \underline{47.78} & 8.52 & \underline{9.38}
& 76.50 & \underline{47.11} & 29.65 \\

Relay-OPD
& 17.29 & 16.67 & \textbf{15.31}
& 46.99 & \textbf{12.12} & 7.29
& \underline{76.70} & 45.33 & \underline{29.71} \\

MAESTRO
& \underline{17.50} & \textbf{21.98} & \underline{13.96}
& \textbf{50.60} & \underline{11.46} & \textbf{11.35}
& \textbf{76.81} & \textbf{47.22} & \textbf{31.36} \\

\midrule
\multicolumn{10}{c}{\textit{Qwen3-1.7B-Non-Thinking}} \\
\midrule

Student
& 12.60 & 9.58 & 7.40
& 47.89 & 6.34 & 4.38
& 71.95 & 38.54 & 24.84 \\

SFT
& 23.33 & 19.48 & 16.15
& 59.45 & 12.59 & 6.56
& 81.40 & 46.62 & 33.20 \\

KD
& 23.54 & 21.15 & 15.31
& 60.23 & 12.78 & 7.50
& 81.45 & 48.07 & 33.75 \\

GRPO
& 24.58 & 22.08 & 15.62
& 60.16 & 14.49 & 9.38
& 80.35 & 48.74 & 34.42 \\

TRD
& 19.27 & 19.69 & 12.71
& 55.47 & 11.93 & 4.58
& 77.70 & 44.18 & 30.69 \\

OPD
& 35.83 & 25.52 & 23.33
& \underline{70.08} & 20.08 & 14.06
& 85.70 & 55.27 & 41.23 \\

SKD
& 30.00 & 27.29 & 25.31
& 62.24 & 19.98 & 11.35
& 84.90 & 55.70 & 39.60 \\

FastOPD
& 39.38 & 25.31 & 27.29
& 67.47 & 23.67 & \underline{20.63}
& 86.40 & 57.33 & 43.43 \\

Relay-OPD
& \underline{44.69} & \underline{30.00} & \underline{28.02}
& 68.26 & \underline{24.24} & 16.67
& \underline{88.80} & \underline{60.74} & \underline{45.18} \\

MAESTRO
& \textbf{45.10} & \textbf{33.13} & \textbf{30.10}
& \textbf{71.65} & \textbf{25.66} & \textbf{21.15}
& \textbf{88.85} & \textbf{61.30} & \textbf{47.12} \\

\bottomrule
\end{tabular}
}
% \vspace{-2pt}
\end{table*}

\subsection{Main Results}

The results of different methods on eight mathematical reasoning benchmarks are
shown in Table~\ref{tab:main_results}.

First, MAESTRO achieves the best overall performance under both student
settings. In particular, with the Qwen3-1.7B student, MAESTRO achieves the best result on all eight
benchmarks, while with the 0.6B student, it ranks
among the top two across all benchmarks. These results demonstrate the broad
effectiveness of MAESTRO across different benchmarks and student scales.

Second, the improvement is particularly pronounced on challenging competition
problems. On the AIME and HMMT benchmarks, MAESTRO improves over Relay-OPD
by 2.30 percentage points on average with the 1.7B student, and by 7.26 points
over standard OPD. This suggests that adaptive teacher intervention is
especially beneficial for more challenging reasoning problems.

Third, MAESTRO consistently improves upon both standard OPD and existing
teacher-intervention methods. This shows that the benefit does not come from
teacher intervention alone, but from adapting when and how much teacher
guidance is introduced according to the current reasoning state.

\begin{figure*}[t]
\centering

% ==================== (a) Ablation ====================
\begin{minipage}[t]{0.38\textwidth}
\centering
\vspace{0pt}

\small
\setlength{\tabcolsep}{5.0pt}
\renewcommand{\arraystretch}{1.08}
\begin{tabular}{lcc}
\toprule
Variant & Avg. & Pass@3 \\
\midrule
MAESTRO          & \textbf{47.1} & \textbf{59.2} \\
Cov. only        & 43.3 & 55.8 \\
Sim. only        & 44.5 & 55.1 \\
w/o weighted-agg & 43.6 & 54.3 \\
w/o adapt. depth & \underline{45.1} & \underline{56.5} \\
\bottomrule
\end{tabular}

% compensate for the shorter table body
\vspace{12pt}

{\footnotesize\textbf{(a) Ablation Study}}

\end{minipage}
\hfill
% ==================== (b) PDS ====================
\begin{minipage}[t]{0.285\textwidth}
\centering
\vspace{0pt}

\includegraphics[
    width=\linewidth
]{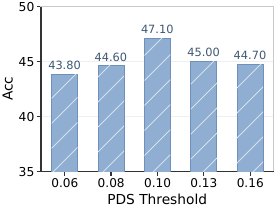}

\vspace{-2pt}

{\footnotesize\textbf{(b) PDS Sensitivity}}

\end{minipage}
\hfill
% ==================== (c) Length ====================
\begin{minipage}[t]{0.295\textwidth}
\centering
\vspace{0pt}

\includegraphics[
    width=\linewidth
]{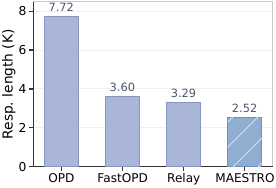}

% C is slightly shorter than B
\vspace{4pt}

{\footnotesize\textbf{(c) Training Response Length}}

\end{minipage}

\vspace{-2pt}

\caption{
Ablation, sensitivity, and training-efficiency analysis of MAESTRO.
(a) Component ablation.
(b) PDS threshold sensitivity.
(c) Average training response length.
% ; MAESTRO uses substantially shorter
% rollouts than OPD, FastOPD, and Relay-OPD
}
\label{fig:ablation_sensitivity_efficiency}

\end{figure*}

\subsection{In-depth Analysis}

\paratitle{Ablation Study.}
We first study the contribution of each component in MAESTRO. As shown in
Fig.~\ref{fig:ablation_sensitivity_efficiency}(a), the full method achieves the best
results in both average accuracy and Pass@3. Using either coverage
(\emph{Cov. only }) or distribution similarity
(\emph{Sim. only}) alone causes a clear performance drop, showing that
the two terms provide complementary information for measuring policy
disagreement. Removing weighted aggregation
(\emph{w/o weighted-agg}) also hurts performance, while using a fixed
intervention span (\emph{w/o adapt. depth}) gives a smaller but consistent drop.
These results support both the design of PDS and the adaptive intervention
strategy.

\paratitle{Sensitivity Analysis.}
We further examine the sensitivity of MAESTRO to the PDS threshold.
As shown in Fig.~\ref{fig:ablation_sensitivity_efficiency}(b), performance
increases as the threshold moves from 0.06 to 0.10 and decreases thereafter,
reaching the best average accuracy of 47.12 at 0.10. This suggests that an
intermediate threshold provides a better balance: a low threshold may trigger
teacher intervention too frequently, while a high threshold may miss useful
interventions.

\paratitle{Efficiency Analysis.}
MAESTRO reduces the average training response length by 67.3\% relative to OPD, leading to substantially lower rollout cost.
This improvement reflects our RQ2 finding (Section~2.2.2) that appropriate
teacher intervention can stabilize training under a limited rollout length
while reducing unnecessary student continuation.
MAESTRO operationalizes this trade-off through adaptive intervention, reducing unnecessary rollout tokens while maintaining effective learning.

\begin{figure}[t]
    \centering
    \includegraphics[width=\linewidth]{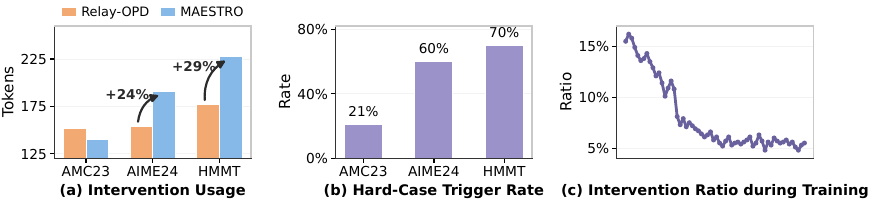}
    \caption{
Intervention behavior across problem difficulty and training.
(a) Average intervention usage per trajectory.
(b) Hard-case trigger rate across the same benchmarks.
(c) Intervention ratio during training, measured as the fraction of
teacher-generated tokens in each response.
}
    \label{fig:takeover_behavior}
    \vspace{-6pt}
\end{figure}

\subsection{Adaptive Intervention Analysis}
\label{sec:adaptive_analysis}

We further examine whether MAESTRO actually exhibits the adaptive behavior
motivated by our analysis. We focus on two aspects: whether teacher guidance
adapts to problem difficulty, and whether intervention changes as the student
improves during training.

\paratitle{Adaptation to Problem Difficulty.}
MAESTRO automatically allocates more teacher guidance to harder problems.
As shown in Fig.~\ref{fig:takeover_behavior}(a), it uses fewer teacher tokens
than Relay-OPD on the easier AMC23, but increases teacher usage on the harder
AIME24 and HMMT by 24\% and 29\%, respectively. This adaptation is also
reflected in intervention depth. Fig.~\ref{fig:takeover_behavior}(b) shows
that the hard-case trigger rate rises from 21\% on AMC23 to 60\% on AIME24
and 70\% on HMMT. These results show that MAESTRO uses PDS to adapt not only whether to
intervene, but also how much teacher guidance to provide according to the
difficulty of the current trajectory.

\paratitle{Adaptation During Training.}
MAESTRO also adjusts its intervention behavior as the student improves.
As shown in Fig.~\ref{fig:takeover_behavior}(c), the overall intervention ratio
gradually decreases from about 16\% to around 5\% during training. Rather than
maintaining a fixed level of teacher guidance, MAESTRO progressively returns
more of the rollout to the student as the two policies become closer. This
shows that its intervention naturally adapts to the evolving student policy.

\section{Conclusion}

Teacher intervention in OPD is not simply a matter of providing more teacher
guidance. Our analysis reveals a structured interaction between intervention
and student learning. Rollout quality and off-policy load both scale
approximately logarithmically with intervention depth, while moderate
intervention raises peak performance and stronger intervention improves
retention. Moreover, the preferred intervention depth and position vary with
problem difficulty and reasoning progress. Together, these findings suggest
that effective teacher guidance should be selective and adaptive rather than
uniformly applied.

Motivated by these observations, we introduce \textbf{MAESTRO}, which uses
local policy disagreement to adapt both the timing and depth of teacher
intervention. Across eight mathematical reasoning benchmarks and two student
scales, MAESTRO achieves the highest macro-average accuracy among the compared
methods while reducing average training response length by 67.3\% relative to
standard OPD. More broadly, our results suggest that the key to effective
OPD is not maximizing teacher involvement, but orchestrating
teacher guidance around where and how much the student needs it.

\bibliographystyle{iclr2027_conference}
\bibliography{iclr2027_conference}

\appendix
\section{Related Work}

\paratitle{On-Policy Distillation.}
On-policy distillation (OPD) trains the student on trajectories sampled from
its current policy, allowing teacher supervision to remain aligned with the
states encountered by the student~\citep{DBLP:conf/iclr/AgarwalVZSGGB24}.
Recent studies have begun to examine the mechanisms and limitations of OPD.
Rethinking OPD identifies teacher--student compatibility and informative
teacher knowledge as key factors for successful distillation, and characterizes
training as progressive alignment on high-probability tokens~\citep{DBLP:journals/corr/abs-2604-13016}.
FastOPD further finds that useful supervision is concentrated toward reasoning
prefixes, showing that shorter rollouts can retain much of the benefit of full
OPD at substantially lower training cost~\citep{DBLP:conf/acl/ZhangYJHRQLBT26}.
Beyond standard OPD, recent work also explores trajectory-level correction.
TRD refines problematic student trajectories with teacher guidance, while
Relay-OPD briefly hands generation to the teacher at detected failure points
before returning control to the student~\citep{DBLP:journals/corr/abs-2606-08432,DBLP:journals/corr/abs-2607-26057}.
These studies motivate a closer examination of how much teacher guidance
should be introduced and how it should be allocated along student trajectories.

\paratitle{Reasoning Post-Training.}
Recent advances in mathematical reasoning have shown that substantial reasoning
capabilities can be elicited through post-training. DeepSeek-R1 demonstrates
that large-scale reinforcement learning can induce long-form reasoning
behaviors, while subsequent work has improved the effectiveness and
reproducibility of reinforcement learning with verifiable rewards~\citep{DBLP:journals/corr/abs-2501-12948,DBLP:conf/nips/YuZZYZYDFLLLLLL25}.
Other approaches improve mathematical reasoning through iterative
self-improvement, reward-guided training, and search-based reasoning~\citep{DBLP:journals/corr/abs-2409-12122,DBLP:conf/icml/GuanZLSSZ0025}.
These advances establish mathematical reasoning as an important setting for
studying long-horizon model improvement. In contrast to reward-based
optimization, our work focuses on how dense teacher supervision can be
selectively introduced into student reasoning trajectories.
% \appendix

\section{Analysis Experiment Details}
\label{app:analysis_details}

We provide additional details for the controlled experiments in
Section~\ref{sec:analysis}. Unless otherwise specified, the analysis
experiments use the same rollout hyperparameters as the main experiments and
final evaluation. Unless otherwise specified, each reported evaluation result is computed with avg@32.

\subsection{RQ1: Scaling Trade-off of Teacher Intervention}

For RQ1, we use a Qwen3-0.6B checkpoint obtained after 35 steps of OPD
training. The trained checkpoint provides sufficiently long reasoning
trajectories for controlled intervention analysis. Starting from the same
checkpoint, we systematically vary the intervention depth while keeping the
evaluated problems and rollout configuration fixed. For each configuration,
we measure both trajectory accuracy and the trajectory-normalized off-policy
load defined in Section~\ref{sec:analysis}. This controlled comparison
isolates how increasing teacher intervention affects trajectory quality and
off-policy deviation.

\subsection{RQ2: Intervention for Stable and Efficient Training}

\paratitle{Intervention under a limited rollout length.}
We use Qwen3-0.6B as the student and deliberately restrict the maximum response
length to 515 tokens. We vary the amount of teacher intervention while keeping
the remaining training configuration fixed. This constrained setting makes
the performance boundary observable within a limited rollout length, allowing
us to compare how different levels of teacher intervention affect the
attainable performance peak and subsequent training stability.

\paratitle{Post-Intervention Student Continuation.}
We further examine how much student rollout is necessary after teacher
intervention. We set the PDS threshold to 0.14 and allow at most two teacher
interventions per trajectory. After the final intervention, we vary the
student continuation budget among 0, 1K, 2K, and 4K tokens, where a budget of
0 terminates the trajectory immediately after the final teacher intervention.
All other training and rollout settings are kept fixed. We compare the
resulting performance, response length, rollout cost, and training efficiency.

\subsection{RQ3: Adaptive Allocation of Teacher Intervention}

For RQ3, we use the same Qwen3-0.6B checkpoint after 35 steps of OPD training
as in RQ1. We conduct controlled experiments over intervention depth and
relative intervention position while keeping the model checkpoint and rollout
configuration fixed. For intervention depth, we compare problems of different
difficulty to study how the preferred amount of teacher guidance changes
across trajectories. For intervention position, we vary where teacher
intervention occurs along the reasoning trajectory and measure the resulting
trajectory quality.

\paratitle{Normalized Improvement.}
To compare intervention effects across problem groups with different baseline
difficulty and attainable accuracy ranges, we normalize the observed accuracy
within the available improvement range of each group. For problem group $g$,
we define
\begin{equation}
    I_{g,t}
    =
    \frac{A_{g,t}-A_{g,\mathrm{base}}}
         {A_{g,\mathrm{peak}}-A_{g,\mathrm{base}}},
\end{equation}
where $A_{g,t}$ denotes the accuracy under the current intervention
configuration, $A_{g,\mathrm{base}}$ is the minimum accuracy before teacher
intervention, and $A_{g,\mathrm{peak}}$ is the maximum accuracy reached after
teacher intervention. Under this normalization, $I_{g,t}=0$ corresponds to
the pre-intervention minimum and $I_{g,t}=1$ corresponds to the
post-intervention peak. This normalization accounts for differences in both
absolute accuracy and available improvement range, allowing the relative
benefit of teacher intervention to be compared across problem groups of
different difficulty.

\section{Training Details}
\label{app:training_details}

We provide additional implementation details for the main experiments. For
training hyperparameters shared with Relay-OPD, we follow its official
implementation.

\paratitle{Optimization.}
We use a global batch size of 128 and a PPO mini-batch size of 128, with one
optimization epoch per batch. The learning rate is fixed at
$1\times10^{-6}$ throughout training, and each prompt uses one rollout during
online distillation. All methods are trained for one epoch. The maximum prompt
and response lengths are 2,048 and 8,192 tokens, respectively. Training is
conducted on eight NVIDIA H20 GPUs, with four GPUs allocated to the student
and four to the teacher.

\paratitle{Rollout and Evaluation.}
Trajectory generation uses a sampling temperature of 1.0 and top-$p$ of 1.0.
The same sampling hyperparameters are used for the analysis experiments, main
training experiments, and final evaluation. During evaluation, we
independently sample 32 responses for each problem and report avg@32 accuracy.

\paratitle{PDS Configuration.}
We set the PDS threshold to 0.14 for Qwen3-0.6B and 0.10 for Qwen3-1.7B.
The smaller student exhibits a larger inherent policy gap from the teacher and
therefore uses a slightly higher intervention threshold. For both model scales,
we allow at most two teacher interventions per trajectory.
PDS is aggregated over natural reasoning paragraphs rather than a fixed
token-level sliding window. We identify paragraph boundaries in the generated
reasoning trajectory and make intervention decisions at paragraph boundaries.
Within each paragraph, we use the weighted aggregation defined in
Section~\ref{sec:pds}, with $\alpha=0.5$ and $\tau=12$. This design assigns
greater importance to disagreement earlier in a reasoning paragraph while
aligning intervention decisions with coherent reasoning segments.

We cap the number of teacher interventions at 2 primarily to control rollout
efficiency and maintain a comparable intervention regime to Relay-OPD.
Because the rollout terminates after the final intervention, allowing additional
interventions also permits longer student--teacher interaction before
termination and can increase rollout cost. We therefore treat the intervention
cap as an efficiency constraint rather than a performance-tuned hyperparameter.

\section{Additional Analysis and Robustness}
\label{sec:additional_analysis}

\subsection{Sensitivity to Aggregation Hyperparameters}
\label{sec:aggregation_sensitivity}

\begin{figure}[t]
    \centering
    \includegraphics[width=0.60\linewidth]{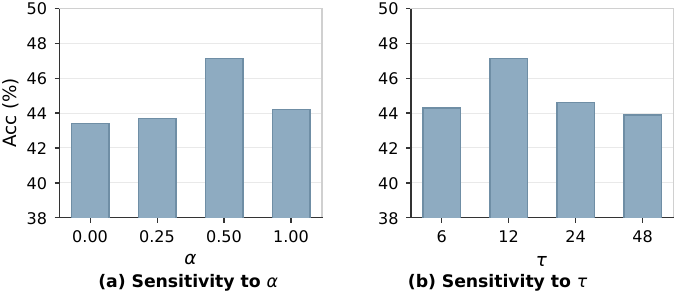}
    \caption{
    Sensitivity of MAESTRO to the paragraph-level aggregation hyperparameters.
    (a) Sensitivity to the weighting strength $\alpha$.
    (b) Sensitivity to the decay scale $\tau$.
    The default configuration, $\alpha=0.5$ and $\tau=12$, achieves the
    highest accuracy in both sweeps.
    }
    \label{fig:aggregation_sensitivity}
\end{figure}

MAESTRO uses two hyperparameters in paragraph-level PDS aggregation:
$\alpha$ controls the additional weight assigned to earlier tokens, while
$\tau$ controls how quickly this emphasis decays within a reasoning paragraph.
To examine whether the aggregation mechanism depends critically on these
choices, we vary one hyperparameter at a time while keeping all other settings
fixed. The sensitivity analyses in
Figures~\ref{fig:ablation_sensitivity_efficiency}(b)
and~\ref{fig:aggregation_sensitivity}
are conducted with the Qwen3-1.7B-Non-Thinking student.

As shown in Figure~\ref{fig:aggregation_sensitivity}(a), the default
$\alpha=0.5$ achieves an accuracy of 47.1, compared with 43.4, 43.7,
and 44.2 for $\alpha=0$, $0.25$, and $1.0$, respectively.
A similar pattern is observed for $\tau$ in
Figure~\ref{fig:aggregation_sensitivity}(b): $\tau=12$ achieves 47.1,
while $\tau=6$, $24$, and $48$ obtain 44.3, 44.6, and 43.9,
respectively.

These results reveal a clear trade-off in paragraph-level aggregation.
If the positional bias is too weak, the aggregation approaches uniform
averaging and underuses the fact that disagreement near the beginning of a
reasoning paragraph is often more indicative of its subsequent direction.
Conversely, an excessively large weighting strength can make a few early
tokens dominate the decision, while a very small decay scale concentrates the
extra weight too narrowly near the paragraph beginning. A very large decay scale instead spreads the positional bias too broadly,
weakening the distinction between early directional signals and later
continuation. The default setting therefore balances early emphasis with
paragraph-level evidence.

\subsection{Teacher Usage and Intervention Allocation}
\label{sec:teacher_usage}

\begin{figure}[t]
    \centering
    \includegraphics[width=0.98\linewidth]{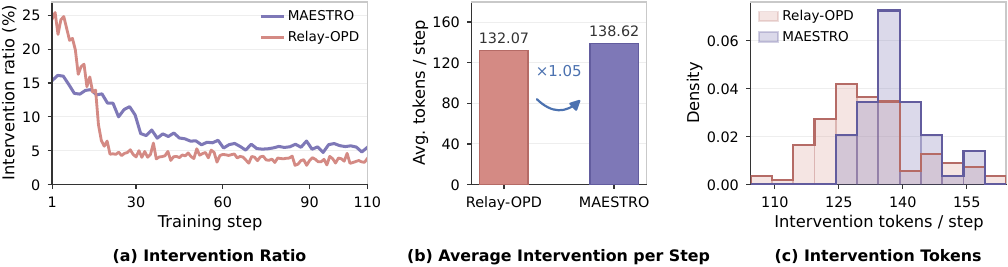}
    \caption{
    Comparison of teacher usage between MAESTRO and Relay-OPD.
    (a) Intervention ratio throughout training.
    (b) Average teacher-token usage per training step, showing comparable overall
    teacher budgets between the two methods.
    (c) Distribution of teacher-token volume across training steps.
    Although their aggregate usage is similar, MAESTRO exhibits a different
    temporal allocation of teacher guidance.
    }
    \label{fig:teacher_usage}
\end{figure}

We further examine whether the gains of MAESTRO can be attributed to simply
using more teacher generation. Figure~\ref{fig:teacher_usage} compares its
intervention behavior with Relay-OPD from both temporal and aggregate
perspectives.

As shown in Figure~\ref{fig:teacher_usage}(a), the two methods exhibit
substantially different intervention dynamics. Relay-OPD concentrates much of
its teacher involvement at the beginning of training and rapidly reduces its
intervention ratio thereafter. In contrast, MAESTRO starts with a lower intervention
ratio and decreases more gradually as training progresses. Despite this
different allocation pattern, their overall teacher usage remains comparable:
Figure~\ref{fig:teacher_usage}(b) shows that MAESTRO uses only $1.05\times$
the average teacher tokens per training step of Relay-OPD. Their token-volume
distributions also largely overlap, as shown in
Figure~\ref{fig:teacher_usage}(c), while MAESTRO concentrates teacher usage
more consistently around a moderate range.

Importantly, teacher generation should not be regarded as a monotonic
performance resource in OPD. As shown in RQ2, increasing intervention does not
continuously improve student performance: moderate intervention raises the
attainable performance peak, whereas excessive intervention can reduce it.
Together with the increasing off-policy deviation identified in RQ1, this
indicates that additional teacher involvement brings both corrective benefit
and distributional cost. Effective intervention therefore requires careful
allocation rather than simply maximizing or exactly matching the number of
teacher-generated tokens.

These results suggest that MAESTRO's improvement is primarily associated with
how teacher guidance is allocated over training, rather than a substantial
increase in the overall teacher budget. By adapting intervention to the
current policy disagreement, MAESTRO redistributes a comparable amount of
teacher guidance toward the states where stronger correction is needed.

\end{document}